# GENDER RECOGNITION BASED ON SIFT FEATURES

Sahar Yousefi, sahar_yousefi@ymail.com

Morteza Zahedi, zahedi@shahroodut.ac.ir

School of Technology and Computer Engineering, Shahrood University of Technology, Shahrood, Iran

This paper proposes a robust approach for face detection and gender classification in color images. Previous researches about gender recognition suppose an expensive computational and time-consuming pre-processing step in order to alignment in which face images are aligned so that facial landmarks like eyes, nose, lips, chin are placed in uniform locations in image. In this paper, a novel technique based on mathematical analysis is represented in three stages that eliminates alignment step. First, a new color based face detection method is represented with a better result and more robustness in complex backgrounds. Next, the features which are invariant to affine transformations are extracted from each face using scale invariant feature transform (SIFT) method. To evaluate the performance of the proposed algorithm, experiments have been conducted by employing a SVM classifier on a database of face images which contains 500 images from distinct people with equal ratio of male and female.

## 1. Introduction

Automatic gender recognition problem has a high application potential in some places where people should be served depend on their gender, such as costumer statistics collection in special places like park places, supermarkets, restaurants and also security surveillance in building entrances. Hence, it is an important problem in computer aided systems which interact with human being.

Several approaches have been proposed to effectively segment and recognize human genders. But most efforts in this field are not robust to some changes like rotation and illumination. In this work, thanks to using Scale Invariant Feature Transform (SIFT) algorithm a novel method for gender recognition is proposed which has a strong adaption to misalignment and affine transformation. An affine transformation is any transformation that preserves co-linearity relation between points and ratios of distances along a line [1,2], like changes of light, rotation, etc. SIFT is a mathematical algorithm for extracting interest point features from images that can be used to perform reliable matching between different views of objects.

Necessary pre-processing step prior to classification into male and female group is regions of interest (ROIs) segmentation. Gender classification is required to delineate the boundaries of the ROIs, ensuring that faces are outlined. In this paper, a new color based segmentation method is used which experiments on complicated images with many details have shown high performance.

The next step is extracting the invariant features and classification of the persons' images into the two categories of male and female. There are some basic differences between male and female faces in hairline, forehead, eyebrows, nose, cheeks, chin, etc [3]. These characteristics are similar in members of one category and different from members of the other one which can be extracted by SIFT algorithm finding the interest points from images that can be used to perform reliable recognition.

Finally, a benchmark database of face images for a group of people was employed by a SVM classifier. The database contains 500 images from distinct people with equal ratio of male and female. The proposed algorithm was performed ten times and in each trial the database was divided in two parts randomly, one part for training set and the other one for testing.

## 2. Related Works

The first attempt to solve gender classification problem was based on neural network and has been done by Cottrell et al. They proposed a two layer neural network, one for face compression and the other layer in order to face classification. The result of hidden layer of compression layer was reduced dimensionally similar to Eigenface method. Such a network indicates 63% accuracy for a database containing 64 photos [4]. In a similar method a two-layer fully connected neural network named SEXNET without dimensionally reduction is used and results reported 91.9% accuracy for a database containing 90 images [5]. This method was followed for a larger database and reported 90% accuracy percentage [6]. After that a multi-layer neural network to identify gender from face images of different solutions is used [7].

In another paper a mixture of radial basis function (RBF) network and decision tree for gender classification was used [8]. Also, Radial Basis Function (RBF) and Perceptron networks for comparing performance of these networks on raw image with PCA-based representations was used and the best performance of 91.8% has been reported for a database contains 160 facial images [9]. Another article is employed support vector machine (SVM) that the result indicates 96.6% accuracy rate with 1755 images by using RBF kernel [10]. Another literature used the wholistic features which extracted by independent component analysis (ICA) and did classification by using linear discriminate analysis (LDA) [11]. In [12] Gaussian process classifiers (GPCs) which are Bayesian kernel classifiers over SVM are introduced for gender classification which this method improved the results of SVM method. On the other hand, [13] used pixel-pattern-based texture feature (PPBTF) for real time gender recognition. In this reference Adaboost method used to select the most discriminative feature subset and support vector machine (SVM) for classification. Ultimately [14] considered the problem of gender classification from frontal facial image using genetic feature subset selection and compared some methods like eigenvector, Bayes, neural network, LDA and SVM for a database contains 400 frontal images from 400 distinct people.

The most literatures about gender recognition have two constraints. First, the database images merely should be containing face region without additional details i.e. the face part in the picture has not to be located in front of a complex background with various colors and regions. The second, the proposed methods so far have an expensive computation step for aligning the faces on a prototype model. The novel method proposed here, not only can detect the faces locating in the pictures with high complexity of backgrounds, but also eliminates the aligning step by using the SIFT features which are invariant to affine transformation.

## 3. Gender Classification

There are some basic differences between male and female facial landmarks which can be used for classification. Generally these characteristics could be applied for all ethnic groups [3]. So gender could be distinguished by people characteristics. In order to find similar male and female characteristics and matching them for classification, the SIFT method is used. Figure 1 presents an overview for our gender classification system. This schema has three significant stages containing:

- Color based face detection
- SIFT feature extraction
- SVM classification

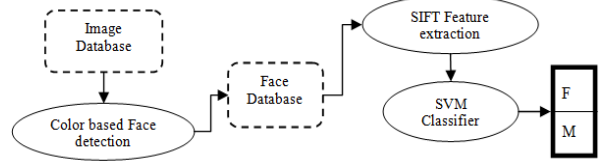

Fig. 1. The process of gender classification

### 3.1. Color based face detection

The face detection contains two stages. First stage is finding region of interest (ROI) that is possibly contains faces. The next stage is determining if the detected ROIs contain face or not.

First, HSV color map for finding skin regions from image was employed. This color space is closer than RGB system to human color perception system. The conversion from RGB to HSV is performed using "Eq.(1)".

$$Hue = \begin{cases} \theta, & if B \leq G \\ 360 - \theta, & if B > G \end{cases} \quad (1)$$

$$Saturation = 1 - \frac{3}{(R+G+B)}[min(R,G,B)]$$

$$Value = \frac{1}{3}(R+G+B)$$

In which, the $\theta$ is defined by "Eq.(2)".

$$\theta = cos^{-1}\left\{\frac{\frac{1}{2}[(R-G)+(R-B)]}{[(R-G)^2+(R-B)(G-B)]^{1/2}}\right\} \quad (2)$$

As figure 2 shows there are some equations between HSV color map parameters (Hue, Saturation, Value) in skin regions.

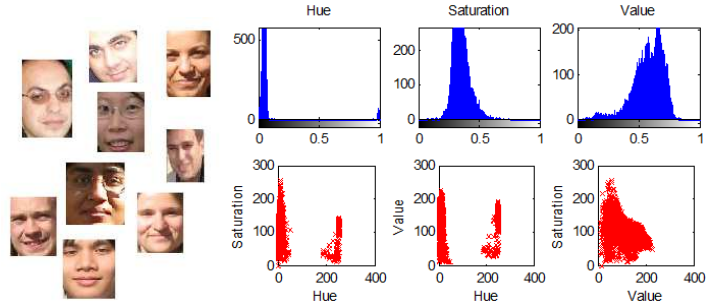

Fig. 2. Relation between HSV parameter in skin regions

Here, a new relation between HSV parameters for finding ROIs was defined. For skin segmentation pixel set that is shown in "Eq.(3)" was found by experience.

$$ROI_{Face} = Hue_{Set} - Saturation_{Set} - Value_{Set} \quad (3)$$

$$Hue_{Set} = \{Region \mid 20 < Hue_{Region} < 200\}$$

$$Saturation_{Set} = \{Region \mid 30 < Saturation_{Region} < 160\}$$

$$Value_{Set} = \{Region \mid 150 < Value_{Region} < 255\}$$

Figure 3 shows the result after finding ROIs for some images. Experimental results demonstrate that this technique is a good skin detection method in complicated images.

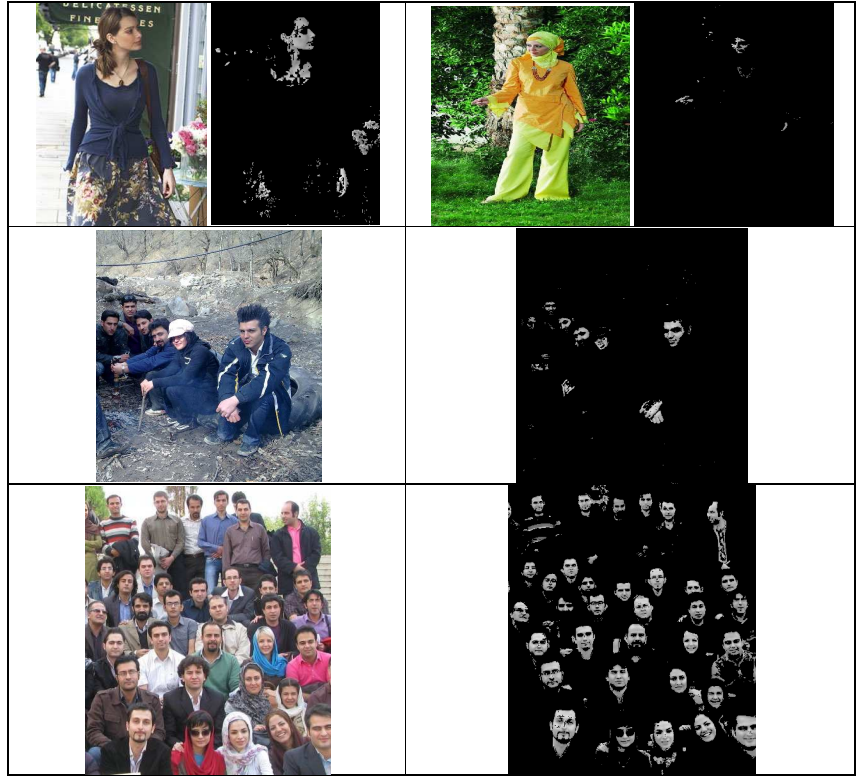
Fig. 3. ROI segmentation using HSV color system

Next, template matching method, for face determining was used. In this technique a face template is convolved with ROI regions that found from previous step. The face template is mean of some training face images (figure 4).

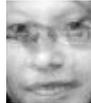
Fig. 4. A Template for face detection

### 3.2. *SIFT feature extraction*
SIFT (Scale invariant feature transform) is a mathematical feature extraction method proposed by David Lowe [15]. The extracted features are invariant to any affine transformation. Under affine transformations, angles, lengths and shapes are not preserved, but the following are some of the properties of a plane figure that remain invariant [15, 16]:
- The concurrency and colinearity of corresponding lines and points;
- The ratio in which a corresponding point divides a corresponding segment.

Extracting features is performed by a cascade filtering approach using a four-stage algorithm.

#### 3.2.1. *Scale-space extrema detection*
First a scale space is defined by "Eq.(4)".

$$L(x, y, \sigma) = G(x, y, \sigma) * I(x, y) \qquad (4)$$

Where * is the convolution operator, $G(x, y, \sigma)$ is a variable-scale Gaussian and $I(x, y)$ is the input image. For finding stable features difference-of-Gaussian function convolved with image, which can be computed with difference of two nearby scales, After each octave, the Gaussian image is down-sampled by a factor of 2, and the process repeated.

$$L(x, y, \sigma) = G(x, y, \sigma) * I(x, y) \qquad (5)$$

To detect the local maxima and minima of $D(x, y, \sigma)$ each point is compared to its eight neighbors at the same scale, plus the nine corresponding neighbors at neighboring scales. If the pixel is a local maximum or minimum, it is selected as a candidate keypoint.

#### 3.2.2. *Keypoint localization and filtering*
This step attempts to eliminate keypoints which are located on edges or the contrast between point its neighbors and is low.

### 3.2.3. Orientation assignment

In this stage, one or more orientation is assigned to each keypoint in order to make them invariant to rotation. Suppose for a keypoint, L is the image with the closest scale, gradient magnitude and orientation can be computed using "Eq.(6)".

$$Gradient vector = \begin{bmatrix} L(x+1,y) - L(x-1,y) \\ L(x,y+1) - L(x,y-1) \end{bmatrix} \quad (6)$$

$$m(x,y) = \sqrt{(L(x+1,y) - L(x-1,y))^2 + (L(x,y+1) - L(x,y-1))^2}$$

$$\theta(x,y) = \tan^{-1}((L(x,y+1) - L(x,y-1))/(L(x+1,y) - L(x-1,y)))$$

In next stage a gradient histogram (36 bins) is created. Any peak within 80% of the highest peak is used to create a keypoint with that orientation.

### 3.2.4. Keypoint descriptor

The final step is to compute a descriptor to make it invariant to remaining variations. For this purpose a 16×16 Gradient window is taken that partitioned into 4×4 sub windows. Then, histogram of 4×4 samples in eight bins is created. This result in a feature vector containing 128 elements.

When at least three keys agree on the model parameters with low residual, there is strong evidence for the presence of the object. Since there may be dozens of SIFT keys in the image of a typical object, it is possible to have substantial levels of occlusion in the image and yet retain high levels of reliability [15].

Whereas there are some basic differences between male and female faces like hairlines, foreheads, noses, eyes, etc, SIFT features are different in two classes and these features could be used for gender classification.

Figure5 shows and two female two male photos with keypoint vectors which SIFT algorithm finds. Most of the features appear on the eyes, nose, mouth and cheeks. As before was mentioned, most similarities appear on the forehead, eyes, eyebrows and the top of nose.

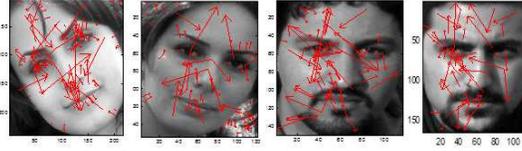

Fig. 5. Keypoint vectors are drawn on the two female and two male's face images, numbers of keypoints are (left to right): 123, 72, 56, 46 respectively.

### 3.3. SVM classifier

Support vector machine (SVM) is a supervised learning technique for pattern classification and regression. SVM is inspired from statistical learning theory and based on concept of structural risk minimization.

For a given training set $\{(x_i, y_i) | x_i \in R^p, y_i \in \{-1,1\}\}_{i=1}^n$ where $y_i$ indicates the class to which sample $x_i$ is belongs. Each $x_i$ is a p-dimensional real vector. The problem is finding the optimal hyper plane that classifies the samples in two classes. The hyper plane is defined in "Eq.(7)".

$$f(x) = \sum_{i=1}^{l} y_i \alpha_i . k(x, x_i) + b \quad (7)$$

Where $\alpha$ and b are constants and $k(.,.)$ is a kernel function and the sign of $f(x)$ determines the class label of sample x. For linear SVM the kernel function is dot product of two N-dimensional vectors "Eq.(8)".

$$k(x_i, x_j) = x_i . x_j \quad (8)$$

While for non linear SVM, kernel function projects the samples to an Euclidean feature space of higher dimensional M via a nonlinear mapping function $\psi : R^N \to F^M, M \gg N$ and construct a hyper plane in F. In this case, kernel function defined as "Eq.(9)", where $\psi$ is the nonlinear projection function.

$$k(x, x_j) = \psi(x) . \psi(x_j) \quad (8)$$

### 4. Database

Although, there are several databases available for face recognition, the researchers in this field face to the problem of lacking of available databases for the gender recognition purpose. Often, in those databases that are available, person's photo has been repeated, e.g. AT&T face database containing 400 images for 40 people with 10 images per person [17]. In gender recognition problem for purpose of avoiding repetition of patterns in training, validation and test parts should be selected one image per person. If we do this in AT&T database, images are reduced to 40 individual images. Hence, we have been collected a database contains 500 images from 500 distinct people with different facial expression and under different lighting conditions and equal rate of males and females in RGB format.

### 5. Experiment results

In this paper, a new approach based on SIFT features was proposed for gender recognition. The whole process of gender classification can be explained in face detection, feature extraction and classification steps.

For examining the proposed method, the collected database was used. The database contains male and female's images, with same proportion. Images were preprocessed and cropped by a novel color face detection method. For this goal skin regions were segmented using HSV color space parameters and then faces were detected using a face template which made by averaging training faces. Then, features were extracted from all the people images in the database using SIFT algorithm. For each image face 150 more frequent SIFT keypoints. Each keypoint descriptor contains 128 attributes that describe that region in a scale and orientation invariant way. SIFT keypoints can be seem as the fingerprint of images, which each fingerprint identifies a unique feature of an image, hence enables us to discover similar features across different images. Ultimately, the extracted SIFT features were rendered to a SVM classifier for purpose of gender classification. There are distinctive differences between male and female faces in forehead, cheek, lips, eyes, etc. Therefore, recognition is based on these differences. In other words, differences between male and female faces and the keypoints which were extracted from faces for classifying were used. The novel approach was performed ten times and in each repetition the database was divided randomly in two parts, 90% for training and 10% for testing. The gender classification was performed using some kernel functions (linear, Quadratic, RBF). Table1 summarized the experiment results.

Table 1. Accurate percentage of experiments

| Kernel classifier | Accuracy percent | Variance |
|---|---|---|
| Linear kernel | 81.15 | 6.2142 |
| Quadratic kernel | 79.09 | 6.3506 |
| RBF kernel | **86.842** | 5.1897 |

## 6. Conclusion

This paper demonstrates a novel technique for face detection based on color space and gender recognition based on SIFT features. The proposed face detection technique with ability of detecting the faces locating in complex backgrounds and the SIFT features with ability of finding the interest points of the faces describing the discriminate characteristics of male and female groups, lead us to construct a system for gender recognition with high robustness in input images and accuracy of recognition output. By employing a nonlinear SVM classifier the proposed method yields a recognition rate of 87% which is appreciable comparing to the other methods with constraints of having simple backgrounds and being time consuming.